\documentclass[11pt,a4paper]{article}

\usepackage[T1]{fontenc}
\usepackage[utf8]{inputenc}
\usepackage[english]{babel}

\usepackage{amsmath,amssymb,amsfonts,mathtools}
\usepackage{geometry}
\usepackage{microtype}
\usepackage{newtxtext}
\usepackage{newtxmath}

\makeatletter
\renewcommand{\maketitle}{%
\begin{flushleft}
{\LARGE\bfseries \@title \par}
\vspace{1em}
{\normalsize \@author \par}
\vspace{1em}
\end{flushleft}
}
\makeatother

\title{From Performance to Representational Adequacy:\\
A Representational Bootstrap Framework for Adaptive Biological Systems}

\author{%
Jacques Raynal$^{1,*}$,
Pierre Slangen$^{2}$,
Elsa Raynal$^{3}$,
Jacques Margerit$^{4}$\\[0.5em]
{\small $^{1}$Laboratory of Bioengineering and Nanosciences (LBN), University of Montpellier, France}\\
{\small $^{2}$EuroMov Digital Health in Motion, University of Montpellier, IMT Mines Al\`es, Al\`es, France}\\
{\small $^{3}$Certified Sophrologist and Dental Assistant, Sensorimotor Practice, Montpellier, France}\\
{\small $^{4}$Emeritus Professor, University of Montpellier, France}\\[0.5em]
{\small $^{*}$Corresponding author: \texttt{raynal.cab@gmail.com}}
}

\date{}

\begin{document}

\maketitle

\begin{abstract}
Observable performance is commonly used to characterize biological systems. In adaptive systems, however, aggregated outputs may remain insufficient for uniquely resolving observational conditions, and richer multivariate representations may themselves retain substantial ambiguity.

This article proposes a representational bootstrap framework for adaptive biological systems. The term bootstrap is used here in a methodological and epistemological sense, not in the statistical resampling sense. New analytical levels emerge progressively when the active representation becomes insufficient for the question under investigation.

The framework is organized around five successive analytical levels: observable performance, conceptual dynamic organization, exploratory multivariate representation, observed longitudinal centroid displacement, and internal approximation of observed longitudinal displacement. These levels are not independent tools, but a progressive sequence through which increasingly adequate analytical representations can be constructed.

The framework is illustrated by three previously reported gait--occlusion studies, used here as a methodological case sequence rather than as new experimental evidence. In the revised first study, neither the aggregated scalar score nor the static exploratory embedding uniquely resolved the occlusal observational probes. The second study therefore shifted the analysis from static representational non-identifiability to longitudinal centroid displacement within a common PCA representation. The third study examined whether this observed representation-dependent transformation could be internally approximated by a supervised model.

The contribution of this article is not a new algorithm, clinical protocol, or dataset. It is the formalization of a bootstrap methodology in which persistent explanatory insufficiency motivates changes in the analytical question and the emergence of new representational levels.
\end{abstract}

\vspace{0.5em}

\section{Introduction}

Biological systems are often evaluated through observable performance. In biomechanics, physiology, rehabilitation, and clinical assessment, measurable outputs such as speed, symmetry, stability, pressure distribution, postural variables, physiological markers, or composite scores are commonly used to characterize the functional state of a system.

The present work addresses a recurrent problem in representation learning: the progressive emergence of analytical levels when an available representation remains insufficient for the question under investigation.

In adaptive biological systems, aggregated observable variables may fail to uniquely resolve experimental or clinical conditions. Introducing a multivariate representation may preserve relationships that are lost through scalar aggregation, but it does not necessarily produce clearly separated or physiologically interpretable states.

Observable analysis remains necessary. Observable variables provide reproducible descriptors, allow comparisons across conditions, and make biological behavior accessible to quantitative analysis. However, scalar aggregation may remain sensitive to preprocessing, weighting, and normalization choices.

A richer multivariate representation may reduce information loss while still leaving substantial overlap between observational conditions. The methodological difficulty is therefore not only that observable performance may fail to reveal organization, but also that an exploratory embedding may remain insufficient for uniquely identifying independently validated condition-specific system states.

This difficulty becomes especially important when a system is observed under constraint. A constraint may not act as a simple causal input producing a linear response. Instead, it may reveal how the system is expressed, reorganized, or compensated under changing conditions. The scientific question may therefore need to shift from identifying a supposedly optimal static configuration to examining how representations change over time and whether those observed changes can be computationally approximated.

The present article proposes a bootstrap framework for addressing this problem. The term bootstrap is used here in a methodological and epistemological sense. It does not primarily refer to statistical resampling, although such methods may be used as auxiliary tools. Rather, it refers to a progressive construction of knowledge in which new analytical levels emerge from the insufficiency of previous ones.

Throughout this article, the term \textit{bootstrap} refers to methodological emergence unless statistical resampling is explicitly mentioned.

From a machine-learning perspective, the framework is positioned as a methodological contribution to representation learning. It does not introduce a new neural-network architecture or optimization algorithm. Rather, it examines how successive analytical levels can emerge when existing representations remain insufficient for the scientific question being asked.

The central principle is the following:

\begin{quote}
\textbf{Theory should emerge from observation rather than be imposed upon it.}
\end{quote}

This formulation should not be interpreted as a rejection of theory. It describes a disciplined process in which new theoretical levels are introduced only when observation demonstrates that the active representation has become insufficient.

In this framework, analysis begins with observable performance. When the scalar representation remains insufficient, a richer multivariate representation is introduced. If that static representation also fails to resolve the observational conditions, the question shifts toward longitudinal change. If longitudinal displacement can only be observed retrospectively, a further level examines whether that observed transformation can be internally approximated.

The proposed sequence includes five analytical levels:

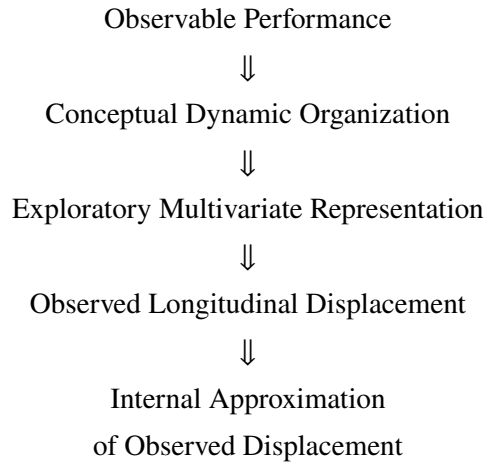
\begin{center}
\renewcommand{\arraystretch}{1.25}
\begin{tabular}{c}
Observable Performance\\
$\Downarrow$\\
Conceptual Dynamic Organization\\
$\Downarrow$\\
Exploratory Multivariate Representation\\
$\Downarrow$\\
Observed Longitudinal Displacement\\
$\Downarrow$\\
Internal Approximation\\
of Observed Displacement
\end{tabular}
\end{center}

These levels are not independent modules and do not imply progressively deeper access to physiological reality. They form a sequence of analytical reformulations generated when the preceding representation remains insufficient.

The three gait--occlusion studies are used as a chronological methodological case sequence. The first study showed persistent static representational non-identifiability. The second introduced longitudinal centroid displacement within a common PCA representation. The third examined whether that observed displacement could be internally approximated within the same single-subject dataset \cite{Raynal2026a,Raynal2026b,Raynal2026c}.

The contribution of the present article is therefore not a new dataset, a new clinical protocol, or a new machine-learning algorithm. Its contribution is the formalization of a quantitative bootstrap methodology for adaptive biological systems. The aim is to show how increasingly adequate representations can emerge from observation when existing levels of analysis become insufficient.

The objective is not to establish a new causal theory of gait or occlusion. Rather, it is to formalize a general methodological framework for studying adaptive systems when complete mechanistic explanations remain unavailable, incomplete, or insufficiently explanatory.

By reframing scientific progress as a sequence of representational transitions, the proposed framework offers a way to study adaptive systems without reducing them to observable performance and without prematurely imposing complete causal models.


\section{The Bootstrap Principle}

The framework proposed in this article originates from a simple observation: adaptive living systems frequently exceed the explanatory capacity of the categories initially used to describe them.

Scientific investigation often begins with a theoretical model. Variables are selected, hypotheses are formulated, and observations are interpreted within a predefined conceptual framework. This approach has generated major advances across numerous scientific disciplines. However, its effectiveness depends on the availability of sufficiently mature explanatory models.

In adaptive biological systems, this condition is not always satisfied.

Living systems are characterized by redundancy, compensation, historical dependence, and continuous reorganization. Multiple internal configurations may generate similar observable outputs, while apparently similar states may diverge over time. Under such conditions, observable performance alone may fail to reveal the organizational properties that govern system behavior.

The bootstrap framework proposes a reversal of the usual sequence between theory and observation. Instead of beginning with a complete explanatory model, it begins with observation itself. New theoretical levels are introduced only when existing levels become insufficient to account for observed phenomena.

Knowledge is therefore constructed progressively.

The process follows four successive steps:

\begin{enumerate}
\item Observe the system under defined constraints.
\item Identify the limits of the current representational level.
\item Introduce a new level capable of explaining the unresolved observations.
\item Test whether the new level reduces the explanatory gap.
\end{enumerate}

The procedure is iterative. Each representational level remains provisional and may itself reveal limitations that require the emergence of a further level. Scientific understanding develops through successive cycles of observation, insufficiency, reconstruction, and reinterpretation.

The resulting process may be summarized schematically as:
\[
\text{Observation}
\rightarrow
\text{Limitation}
\rightarrow
\text{New Representation}
\rightarrow
\text{New Observation}.
\]

Within this framework, theoretical growth is not driven primarily by speculation but by explanatory necessity. A new level is justified only when an observational paradox cannot be adequately resolved within the existing framework.

This principle distinguishes the bootstrap approach from both purely empirical and purely theoretical methodologies.

It is not purely empirical, because each new level introduces conceptual structures that are not directly observable. Low-dimensional representations, longitudinal displacement descriptors, and internal computational approximations are examples of such constructions.

It is not purely theoretical, because each conceptual extension must be justified by a limitation encountered in observation.

The framework therefore establishes a dynamic relationship between empirical evidence and theoretical development.

Theory does not disappear.

Theory emerges.

\subsection{Explanatory Insufficiency as a Scientific Resource}

In conventional scientific reasoning, explanatory insufficiency is often viewed as a problem to be eliminated. Within the bootstrap framework, it acquires a different status: insufficiency becomes informative.

When observations resist explanation, they indicate that the current representational level has reached its limits. The discrepancy between what is observed and what can be explained becomes the source of theoretical innovation.

Scientific progress therefore does not arise solely from confirmation. It also arises from the recognition of inadequacy.

Every bootstrap transition begins with a failure of description.

Observable performance becomes insufficient.

Static multivariate representation remains insufficient.

Retrospective longitudinal displacement remains insufficient to establish internal approximability.

Each insufficiency opens the possibility of a more appropriate analytical representation.

\subsection{Emergence Rather Than Discovery}

The bootstrap framework also modifies the meaning of scientific explanation.

Classical scientific narratives often describe knowledge as the progressive discovery of an underlying reality that already exists independently of observation. The present framework adopts a more modest position.

The objective is not necessarily to discover the final hidden structure of the living system. The objective is to construct representations that account increasingly well for observed behavior.

The relevant question therefore becomes:

\begin{quote}
\textbf{Which representational level best explains the observations currently available?}
\end{quote}

Rather than seeking a definitive explanatory endpoint, the bootstrap framework emphasizes progressive refinement.

Knowledge advances through increasingly adequate representations rather than through access to an ultimate description of reality.

This perspective is particularly relevant for adaptive systems whose organization continuously evolves through interaction with constraints.

\subsection{The Bootstrap Criterion}

The central methodological criterion of the framework can be stated as follows:

\begin{quote}
\textbf{A new theoretical level should be introduced only when observational evidence demonstrates the insufficiency of existing levels.}
\end{quote}

This criterion imposes a form of epistemological discipline. It prevents premature theoretical inflation while encouraging conceptual innovation when empirical observations require it.

Theoretical expansion is therefore constrained by observation.

The framework does not ask:

\begin{quote}
\emph{What theory can be imagined?}
\end{quote}

It asks:

\begin{quote}
\emph{What theory has become necessary?}
\end{quote}

This distinction constitutes the foundation of the bootstrap approach and provides the logical transition toward the five analytical levels described in the following section.

\section{The Emergence of the Five Analytical Levels}

From a machine-learning perspective, the five levels may be interpreted as successive representational layers introduced when the preceding representation no longer accounts for the observed structure of the data.

The bootstrap principle acquires its practical significance only when applied to a concrete scientific investigation. The five analytical levels presented in this framework did not emerge from a predefined theoretical architecture. Rather, they appeared progressively as responses to successive explanatory limitations encountered during the study of an adaptive biological system.

Each level was introduced because the preceding one became insufficient to account for the observed phenomena.

The resulting sequence forms a progressive reconstruction of understanding:

\begin{center}
    \begin{tabular}{c}
    Observable Performance
    $\rightarrow$
    Dynamic Organization
    $\rightarrow$
    Exploratory Multivariate Representation
    \\[0.3em]
    $\rightarrow$
    Observed Longitudinal Viability
    $\rightarrow$
    Internal Approximation of Observed Displacement
    \end{tabular}
    \end{center}

This logic is cumulative rather than substitutive. Each level preserves the insights of the previous one while extending its explanatory capacity.

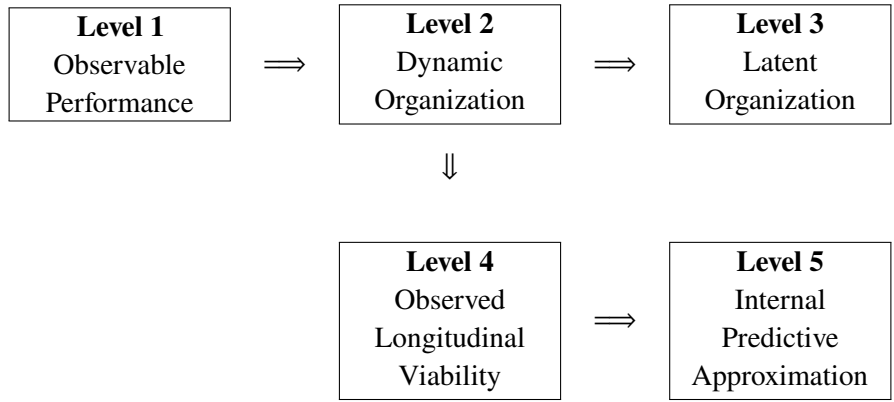
\begin{figure}[htbp]
\centering

\renewcommand{\arraystretch}{1.3}

\begin{tabular}{ccccc}

\fbox{\begin{minipage}{2.7cm}
\centering
\textbf{Level 1}\\
Observable\\
Performance
\end{minipage}}
&
$\Longrightarrow$
&
\fbox{\begin{minipage}{2.7cm}
\centering
\textbf{Level 2}\\
Dynamic\\
Organization
\end{minipage}}
&
$\Longrightarrow$
&
\fbox{\begin{minipage}{2.7cm}
\centering
\textbf{Level 3}\\
Latent\\
Organization
\end{minipage}}

\\[0.7cm]

&& $\Downarrow$ &&

\\[0.7cm]

\multicolumn{2}{c}{}
&
\fbox{\begin{minipage}{2.7cm}
\centering
\textbf{Level 4}\\
Observed\\
Longitudinal Viability
\end{minipage}}
&
$\Longrightarrow$
&
\fbox{\begin{minipage}{2.7cm}
\centering
\textbf{Level 5}\\
Internal Predictive\\
Approximation
\end{minipage}}

\end{tabular}

\vspace{0.6cm}

\begin{tabular}{c}
\textbf{Article 1} \cite{Raynal2026a}
\\
Observable Performance $\rightarrow$ Dynamic Organization $\rightarrow$ Exploratory Multivariate Representation
\\[0.25cm]
\textbf{Article 2} \cite{Raynal2026b}
\\
Exploratory Multivariate Representation $\rightarrow$ Observed Longitudinal Displacement
\\[0.25cm]
\textbf{Article 3} \cite{Raynal2026c}
\\
Observed Longitudinal Displacement $\rightarrow$
Internal Approximation of Observed Displacement
\end{tabular}

\caption{
Bootstrap emergence of successive representational levels generated by explanatory insufficiency. Each analytical level appears when the previous level becomes insufficient to account for the observations. The three previously reported studies successively motivated the transition from observable performance and dynamic organization to exploratory multivariate representation, from exploratory multivariate representation to observed longitudinal displacement, and from observed longitudinal displacement to internal approximation.
}

\label{fig:bootstrap_framework}
\end{figure}

From a machine-learning perspective, the representational bootstrap can be interpreted as a progressive reformulation process in which a new analytical level is introduced whenever the active representation remains insufficient to account for the observed structure of the data.

\subsection{Relationship to the Previous Articles}

The present article does not repeat the empirical analyses developed in the preceding studies. Its objective is to formalize the logic by which those studies progressively generated new analytical questions.

Article~1 showed that the scalar representation was sensitive to its construction and that the exploratory UMAP embedding did not produce independently separated condition-specific clusters. The relevant result was persistent static representational non-identifiability rather than the identification of distinct condition-specific regions within the selected representation \cite{Raynal2026a}.

Article~2 addressed this unresolved static ambiguity by examining M1--M2 centroid displacement within a common PCA representation. It introduced observed longitudinal displacement as a representation-dependent descriptor, while explicitly avoiding its interpretation as a validated physiological viability measure \cite{Raynal2026b}.

Article~3 examined whether this observed coordinate transformation could be internally approximated by a simplified supervised model. Its result concerned internal approximation within the same single-subject dataset, not prospective clinical prediction or recovery of true physiological dynamics \cite{Raynal2026c}.

The present article therefore treats the preceding studies as a sequence of explanatory insufficiencies and analytical reformulations.

\subsection{Level 1: Observable Performance}

The first level corresponds to the traditional language of clinical and biomechanical evaluation.

At this stage, the system is described through observable outputs such as walking speed, cadence, asymmetry indices, postural variables, physiological measurements, or composite performance scores.

These indicators are indispensable because they provide measurable, reproducible, and comparable descriptions of behavior.

However, this level contains an implicit assumption:

\begin{quote}
\textbf{Similar performance implies similar organization.}
\end{quote}

For many engineered systems, this assumption may be acceptable. Adaptive biological systems frequently violate it.

A living system may preserve performance through compensation. Functional outputs can remain stable despite substantial internal reorganization. Conversely, modest observable variations may reflect deeper changes in system organization.

Performance therefore describes what the system produces without necessarily explaining how it produces it.

The first bootstrap transition emerges when performance becomes insufficient to explain observed differences.

\subsection{Level 2: Dynamic Organization}

The second level introduces the notion of organization.

The system is no longer considered a collection of isolated measurements but a coordinated process evolving through time under constraint.

The focus shifts from outputs to the relationships among outputs.

The question changes.

Instead of asking:

\begin{quote}
\emph{How well does the system perform?}
\end{quote}

one asks:

\begin{quote}
\emph{How is the system organized in order to perform?}
\end{quote}

The introduction of dynamic organization acknowledges that living systems are adaptive processes rather than static entities.

The same observable outcome may arise from different regulatory strategies.

However, organization remains only partially accessible through direct observation.

Although temporal analysis reveals more than isolated measurements, it does not fully expose the multidimensional relationships that structure the system.

A further representational level therefore becomes necessary.

\subsection{Level 3: Exploratory Multivariate Representation}

The third level introduces a multivariate embedding in order to preserve relationships that are removed by scalar aggregation.

Its objective is not to reveal a hidden anatomical or physiological reality. It provides a model-dependent representation of relationships among high-dimensional observations.

In the revised founding study, the selected UMAP representation did not reveal clearly separated condition-specific clusters. The six observational probes occupied substantially overlapping regions, and OC2.5 and OC3 remained broadly overlapping \cite{Raynal2026a}.

This negative result is methodologically important. It shows that introducing a richer representation does not necessarily resolve the ambiguity inherited from the scalar level.

Level~3 therefore reduced information loss relative to aggregation, but remained insufficient for uniquely identifying independently validated condition-specific system states.

The next bootstrap transition did not arise because Level~3 had successfully revealed distinct exploratory multivariate representations. It arose because static multivariate representation remained insufficient. The question therefore shifted from where observations were located at one session to how their condition-level representations changed between sessions.

\subsection{Level 4: Observed Longitudinal Displacement}

Level~4 introduces a longitudinal descriptor when static representations remain insufficient.

The selected conditions were projected into a common PCA coordinate system, and the Euclidean displacement between their M1 and M2 centroids was calculated \cite{Raynal2026b}.

This displacement was used as a retrospective, representation-dependent proxy for lower or greater observed longitudinal reorganization. It was not interpreted as direct physiological viability, biological stability, clinical benefit, or therapeutic suitability.

The Level~4 analysis reported the exploratory ordering

\[
d_{\mathrm{OC3}} < d_{\mathrm{ONL}} < d_{\mathrm{OC2.5}}
\]

within the selected PC1--PC2 projection.

However, covariance normalization modified this ordering. The result therefore showed that the operational measure depended partly on the selected geometry and within-condition dispersion.

The explanatory contribution of Level~4 is consequently not the discovery of stable or viable physiological states. It is the introduction of longitudinal representational change as a new analytical property when static representations remain unresolved.

\subsection{Level 5: Internal Approximation of Observed Displacement}

After Level~4, the unresolved issue was no longer whether longitudinal centroid displacement could be observed, but whether the observed M1--M2 coordinate transformation contained enough structure to be approximated within the same single-subject dataset.

This transition motivated Level~5.

Its objective was deliberately restricted: not to predict unseen patients, establish therapeutic recommendations, forecast future outcomes, or infer physiological viability, but to test the internal approximability of an already observed representation-dependent transformation.

Level~5 therefore emerged as a computational continuation of Level~4. It did not replace retrospective displacement analysis. It asked whether the observed displacement pattern could be represented by a simplified supervised model.

In the third study, internal approximation was introduced only in this restricted methodological sense \cite{Raynal2026c}.

The relevant result was that the observed M1--M2 PCA-coordinate transformation could be approximated internally within the same dataset and that the core Euclidean displacement hierarchy was preserved.

Internal approximability does not establish causal explanation, physiological truth, therapeutic superiority, or generalizable prediction. It shows only that the selected transformation possesses sufficient structure to be represented computationally within the available dataset.

\begin{quote}
Level~4 retrospectively describes observed longitudinal centroid displacement.
\end{quote}

\begin{quote}
Level~5 tests whether that observed displacement can be internally approximated.
\end{quote}

\subsection{The Bootstrap Sequence}

Observable performance describes measurable outputs.

Conceptual dynamic organization frames the system as a coordinated adaptive process under constraint.

Exploratory multivariate representation preserves relationships that are removed by scalar aggregation, without necessarily resolving condition identity.

Observed longitudinal displacement describes representation-dependent change between sessions.

Internal approximation tests whether the observed displacement can be represented computationally within the available dataset.

Each level emerges because the preceding representation becomes insufficient for the question under investigation.

The framework therefore does not proceed from theory to observation. It proceeds from observation to increasingly adequate theory.

This progression constitutes the bootstrap logic that underlies the entire framework.

The next section examines the broader epistemological implications of this approach and its relevance for the study of adaptive biological systems.

\section{Bootstrap Epistemology and Adaptive Living Systems}

The five analytical levels described above define more than a methodological framework. They imply a particular conception of scientific knowledge itself.

Classical scientific reasoning often assumes that reality possesses an underlying structure that can progressively be revealed through measurement, experimentation, and theoretical refinement. The objective of science is then understood as the discovery of increasingly accurate descriptions of this hidden structure.

Adaptive living systems challenge this perspective.

Living systems are not static objects awaiting observation. They are dynamic organizations continuously transformed by internal regulation, environmental interaction, historical contingency, and adaptive response. Their observable behavior emerges from processes that remain only partially accessible to direct measurement.

Under these conditions, scientific understanding cannot be reduced to the identification of stable mechanisms alone.

It must also account for emergence, adaptation, compensation, and transformation.

The bootstrap framework responds to this challenge by proposing a different relationship between observation and theory.

Rather than assuming that theory precedes understanding, it considers theory as the result of a progressive reconstruction process.

Knowledge emerges through successive encounters with explanatory insufficiency.

\subsection{Observation as the Origin of Theory}

The central epistemological claim of the bootstrap framework is simple:

\begin{quote}
\textbf{Observation does not merely test theory; observation generates the need for new theory.}
\end{quote}

In many scientific contexts, data are collected in order to confirm or refute pre-existing hypotheses. Within the bootstrap framework, observations play an additional role.

They reveal the limits of the current representational level.

When a phenomenon resists explanation, a new conceptual level becomes necessary.

Theoretical innovation therefore originates not from speculative expansion but from explanatory necessity.

The progression observed in the present framework illustrates this principle.

Performance failed to fully characterize organization.

Organization failed to fully characterize observed longitudinal behavior.

Observed longitudinal displacement remained retrospective and did not determine whether the transformation possessed internally representable structure.

Each failure became the source of a new analytical level.

Scientific progress emerged from the recognition of insufficiency.

\subsection{The Productive Role of Ignorance}

The bootstrap framework assigns a constructive role to ignorance.

Scientific practice often treats ignorance as a temporary absence of knowledge that must eventually be eliminated. Yet in adaptive systems, recognizing what cannot currently be explained may be as important as explaining what is already understood.

An unresolved observation is not merely a deficit.

It is information.

It indicates the boundary of the current explanatory framework.

The bootstrap process therefore transforms ignorance into a methodological resource.

Every explanatory limit becomes a potential source of theoretical emergence.

This perspective encourages intellectual discipline.

A theory is not expanded because it is attractive.

It is expanded because observation requires it.

\subsection{From Discovery to Emergence}

The framework also modifies the meaning of scientific explanation.

Traditional scientific narratives frequently describe knowledge as the progressive discovery of an underlying reality.

The bootstrap perspective adopts a more cautious position.

The objective is not necessarily to uncover a final hidden structure.

The objective is to construct representations that become progressively more adequate to the observations available.

The relevant question is therefore not:

\begin{quote}
\emph{What is the ultimate reality of the system?}
\end{quote}

but:

\begin{quote}
\emph{What representation best accounts for the observed phenomena at the current stage of understanding?}
\end{quote}

This distinction is particularly important when dealing with low-dimensional representations, longitudinal displacement metrics, or internal computational approximations.

These representations should not be interpreted as direct revelations of the living system itself.

They are tools for understanding.

Their value derives from their explanatory usefulness rather than from any claim to ontological finality.

\subsection{Bootstrap and Scientific Inquiry}

The bootstrap framework can be related to the general logic of scientific inquiry.

In Peirce's pragmatist account, scientific knowledge progresses through the iterative correction of hypotheses in response to experience \cite{Peirce1878}. This perspective is useful here only in a methodological sense.

The successive levels proposed in this article should not be interpreted as definitive descriptions of the living system. They are progressively refined representations introduced when previous levels become insufficient to account for the observed data.

In this sense, the bootstrap framework formalizes a self-correcting representational process: observation reveals a limitation, the limitation motivates a new representation, and the new representation is retained only if it reduces the explanatory gap.

\subsection{The Principle of Interpretive Restraint}

A central consequence of the bootstrap framework is the need for interpretive restraint.

As analytical methods become increasingly sophisticated, the temptation arises to confuse representation with reality.

Low-dimensional representations may appear to reveal hidden structures.

Internal approximation models may appear to reveal hidden causes.

Computational regularities may appear to reveal hidden truths.

The bootstrap framework rejects this confusion.

Every representational level remains a model.

Every model remains partial.

Every interpretation remains provisional.

This principle does not weaken scientific inquiry.

On the contrary, it protects scientific inquiry from overinterpretation.

The more powerful the representation becomes, the greater the need for interpretive discipline.

Scientific progress therefore requires not only better observation but also greater discernment.

\subsection{From State-Based Science to Trajectory-Based Science}

The framework suggests a broader transformation in the study of living systems.

Much of traditional science is organized around states.

Variables are measured.

Conditions are compared.

Differences are quantified.

Adaptive systems invite a different perspective.

Their present state cannot be fully understood independently of the trajectory through which it emerged.

Two systems may display similar observable properties while possessing different histories and different capacities for future adaptation.

The bootstrap framework therefore proposes a gradual transition:

\begin{center}
    \begin{tabular}{c}
    Observable Performance
    $\rightarrow$
    Dynamic Organization
    $\rightarrow$
    Exploratory Multivariate Representation
    \\[0.2cm]
    $\rightarrow$
    Observed Longitudinal Viability
    $\rightarrow$
    Internal Approximation of Observed Displacement
    \end{tabular}
    \end{center}

The scientific object progressively changes.

Attention shifts from what the system is to how the system becomes.

This shift does not replace state-based analysis.

It extends it.

The resulting perspective may be relevant not only to biomechanics but also to neuroscience, rehabilitation, physiology, behavioral sciences, and other domains concerned with adaptive living systems.

\subsection{Knowledge as an Adaptive Process}

The deepest implication of the bootstrap framework concerns the nature of scientific knowledge itself.

The adaptive system changes in response to constraints.

The scientific framework changes in response to observations.

In both cases, adaptation occurs through successive reorganizations rather than through immediate access to a complete description.

Knowledge itself becomes a trajectory.

Each explanatory level remains provisional.

Each representation remains revisable.

Each theoretical advance remains open to further emergence.

Scientific understanding is therefore not a destination.

It is an adaptive process of progressively refining the relationship between observation, representation, and explanation.

The next section illustrates this principle through the gait--occlusion studies that originally motivated the development of the bootstrap framework.


\section{The Gait--Occlusion Studies as a Bootstrap Case Study}

The bootstrap framework presented in this article did not originate as a theoretical project. It emerged retrospectively from a sequence of investigations initially designed to explore gait dynamics under controlled occlusal constraints.

These studies did not begin with the objective of constructing a general epistemology of adaptive systems. Their initial purpose was considerably more modest: to determine whether different occlusal configurations could be distinguished through quantitative gait analysis.

Yet the successive investigations revealed a pattern that progressively exceeded the explanatory capacity of the original framework.

Each study generated a question that could not be fully resolved within the analytical level that preceded it.

The bootstrap framework emerged from this progression.

This perspective is consistent with broader approaches to viability theory, biological autonomy, self-organization, and constraint-based adaptive systems \cite{Aubin1991,Varela1979,Kauffman1993,Kelso1995,Newell1986}.

\subsection{First Cycle: The Limits of Observable Performance}

The first investigations focused on observable performance.

Various gait parameters and composite indicators were analyzed under different occlusal conditions. The implicit assumption was straightforward: if one condition produced better performance metrics than another, it could reasonably be considered functionally preferable.

This approach corresponds to the dominant logic of contemporary quantitative evaluation.

Performance is measured.

Conditions are compared.

Differences are interpreted.

However, the observations revealed a limitation.

Configurations displaying relatively similar global performance sometimes appeared to behave differently when examined through a broader analytical lens.

The question emerged naturally:

\begin{quote}
\textbf{Can equivalent performance conceal different forms of organization?}
\end{quote}

This question could not be answered within a purely performance-based framework.

A new representational level became necessary.

The first bootstrap transition had occurred.

\subsection{Second Cycle: The Limits of Static Multivariate Representation}

The introduction of an exploratory multivariate representation transformed the investigation by preserving relationships that were removed through scalar aggregation.

However, the revised Level~3 analysis did not reveal clearly separated condition-specific regions. The six observational probes remained substantially overlapping, and no independently validated condition-specific clusters were identified \cite{Raynal2026a}.

This negative result was methodologically important. It showed that a richer representation does not necessarily resolve the ambiguity inherited from scalar aggregation.

The scientific question therefore shifted. The relevant issue was no longer whether a more complex static representation could reveal discrete states, but whether condition-level representations changed differently between sessions.

Static multivariate representation had reached its explanatory limit. The next bootstrap transition therefore required a longitudinal question.

\subsection{Third Cycle: The Emergence of Observed Longitudinal Displacement}

The next step consisted in introducing longitudinal analysis.

The focus shifted from static location within a selected representation to condition-level change between sessions.

The relevant question became:

\begin{quote}
\textbf{Can observational probes that remain ambiguous in a static representation exhibit different longitudinal centroid displacements?}
\end{quote}

This question led to the introduction of observed longitudinal displacement as an analytical descriptor.

Within the present framework, centroid displacement is not a direct measure of physiological viability or stability. It is a representation-dependent description of how condition-level centroids changed between M1 and M2.

The scientific object changed from static position to observed representational change.

\subsection{Fourth Cycle: The Emergence of Internal Approximation}

The fourth cycle extended the longitudinal question without converting it into clinical prediction.

After Level 4, the unresolved issue was no longer whether longitudinal displacement could be observed, but whether the observed M1--M2 coordinate transformation possessed enough internal structure to be approximated within the same single-subject dataset.

This transition motivated the Level 5 study.

Its objective was deliberately restricted: not to predict unseen patients, not to establish therapeutic recommendations, and not to forecast future clinical outcomes, but to test the internal approximability of an already observed coordinate transformation.

Level 5 therefore emerged as a representational continuation of Level 4. It did not replace retrospective displacement analysis; it asked whether observed longitudinal displacement patterns could become computationally representable.

\subsection{Internal Approximation as a Criterion of Representational Structure}

In the third study \cite{Raynal2026c}, internal approximation was introduced only as a methodological test of representational structure.

From a representation-learning perspective, the importance of this result is that observed coordinate transformations can become computationally representable as structured mappings. The objective is not forecasting but the evaluation of whether the selected representation preserves enough structure to support internal approximation.

The relevant point was not the machine-learning architecture itself, nor the optimization procedure, nor any claim of clinical forecasting. The relevant point was that the observed M1--M2 coordinate transformation could be approximated internally within the same single-subject dataset.

In this restricted sense, internal approximability indicates that the trajectory was not merely a retrospective observation, but a structured transformation that could be represented mathematically.

This does not establish causality, therapeutic superiority, or generalizable prediction. It only justifies the emergence of Level 5 as a new analytical level within the bootstrap sequence.

Within the bootstrap framework, internal approximability becomes a criterion of representational coherence.

A trajectory that can be approximated internally by a simple model is not thereby explained causally. However, within the limits of the available dataset, it is shown to possess enough structure to justify the emergence of a new analytical level.

This interpretation is consistent with the broader use of representation learning as a way to construct informative low-dimensional descriptions of high-dimensional data \cite{Bengio2013}.

Thus, Level 5 does not replace Level 4. It extends it.

Level~4 retrospectively describes observed longitudinal centroid displacement.

Level~5 tests whether observed longitudinal displacement can be represented through internal approximation.

\subsection{From Experimental Sequence to Methodological Framework}

Viewed individually, each study addressed a specific question. Viewed collectively, they reveal a broader methodological pattern.

Article~1 established persistent static representational non-identifiability: neither the scalar score nor the selected exploratory embedding uniquely resolved the observational probes \cite{Raynal2026a}.

Article~2 responded by shifting the analysis toward observed longitudinal centroid displacement within a common PCA representation \cite{Raynal2026b}.

Article~3 examined whether that observed representation-dependent displacement could be internally approximated by a simplified supervised model \cite{Raynal2026c}.

The present article formalizes this sequence as a bootstrap framework.

Each investigation exposed the limitations of the previous analytical level. Each limitation generated the need for a reformulated question and a more appropriate representation.

The progression may be summarized as follows:

\begin{quote}
\textbf{Scalar performance remains insufficiently discriminative.}
\end{quote}

A richer multivariate representation emerges.

\begin{quote}
\textbf{Static multivariate representation remains non-identifying.}
\end{quote}

Longitudinal displacement emerges.

\begin{quote}
\textbf{Retrospective displacement does not establish approximability.}
\end{quote}

Internal approximation emerges.

The resulting sequence was not designed in advance. It was produced by the limitations revealed through observation. This characteristic constitutes the defining feature of the bootstrap framework.

\subsection{Beyond the Particular Case}

The gait--occlusion studies serve here as an illustrative example rather than as the final object of the framework.

The importance of the case study lies not in the specific experimental conditions employed but in the methodological process that it reveals.

The same logic may apply whenever adaptive living systems are investigated under conditions where observable performance proves insufficient to characterize underlying organization.

In such situations, scientific progress may depend less on discovering an immediate explanation than on identifying the limitations of existing representations and allowing new explanatory levels to emerge.

The gait--occlusion studies therefore function as a proof of concept for a broader methodological strategy.

They illustrate how theory can emerge progressively from observation through successive cycles of explanatory insufficiency and representational reconstruction.

The next section examines the broader implications of this framework for scientific reasoning and proposes a general principle for allowing theory to emerge from observation.


\section{Letting Theory Emerge from Observation}

The bootstrap framework proposed in this article is not intended to replace existing scientific methodologies. Experimental protocols, statistical inference, mechanistic modeling, and predictive approaches remain essential components of scientific investigation.

Its contribution lies elsewhere.

The framework provides a methodological orientation for situations in which observations reveal phenomena that exceed the explanatory capacity of existing representations.

In such situations, the central scientific question is no longer simply whether a theory is correct.

The question becomes:

\begin{quote}
\textbf{Which new representational level has become necessary?}
\end{quote}

This shift may appear subtle, yet it fundamentally changes the relationship between observation and explanation.

\subsection{Observation Before Explanation}

Scientific investigation often begins with a model.

A theoretical framework is proposed, variables are selected, hypotheses are formulated, and observations are interpreted through pre-existing conceptual categories.

This strategy has proven remarkably successful whenever the fundamental mechanisms governing a system are already sufficiently understood.

Adaptive living systems present a different challenge.

Their organization is only partially observable. Their behavior emerges from interactions distributed across multiple spatial and temporal scales. Their responses frequently involve compensation, redundancy, and continuous reorganization.

Under such conditions, explanatory categories may become insufficient before their limitations are fully recognized.

The bootstrap framework therefore proposes a methodological inversion.

Observation retains priority.

Theory emerges when observation demonstrates that existing categories no longer provide an adequate account of the phenomena under study.

The objective is not to eliminate theory.

The objective is to allow theory to emerge from explanatory necessity.

\subsection{The Bootstrap Criterion as an Interpretive Rule}

From this perspective follows a fundamental methodological principle:

\begin{quote}
\textbf{A new theoretical level should be introduced only when observation demonstrates the insufficiency of existing levels.}
\end{quote}

This criterion imposes a form of epistemological discipline.

The proliferation of concepts remains constrained by observation.

Theoretical innovation is justified not because it is intellectually attractive, but because it becomes necessary to account for phenomena that remain unexplained.

The bootstrap criterion protects against two symmetrical errors.

The first is reductionism.

Complex phenomena are forced into categories that have become incapable of describing them adequately.

The second is speculative inflation.

New concepts are introduced without a clear observational necessity.

The bootstrap framework seeks a middle path.

Theoretical growth remains possible, but it must be earned through observation.

\subsection{Representation Is Not Reality}

A second consequence concerns the interpretation of scientific models.

As representations become increasingly sophisticated, the temptation arises to regard them as direct descriptions of reality itself.

Representation spaces may appear to reveal hidden structures.

Viability indices may appear to reveal intrinsic properties.

Predictive approximations may appear to reveal invisible causes.

The bootstrap framework calls for continuous interpretive restraint.

Every representational level remains a construction.

Its value lies in its explanatory power rather than in any claim to ultimate truth.

A low-dimensional representation is not the living system.

A centroid-displacement metric is not physiological stability.

An internal approximation is not the future.

These representations are tools that help organize observations.

They should never be confused with the phenomena they seek to describe.

The central risk is not the use of representation, but the loss of distinction between an analytical construct and the biological system it helps to describe.

\subsection{From Certainty to Adequacy}

Classical scientific narratives often suggest a progression toward certainty.

The bootstrap framework proposes a different ideal.

The objective is not certainty.

The objective is adequacy.

A representation becomes valuable when it explains more observations than the representation that preceded it.

Its legitimacy derives from its capacity to reduce explanatory insufficiency.

Scientific progress therefore appears less as a march toward definitive truth than as a succession of increasingly adequate representations.

This perspective is particularly appropriate for adaptive living systems whose organization remains dynamic, context-dependent, and historically situated.

The relevant question is not:

\begin{quote}
\emph{Have we reached the final explanation?}
\end{quote}

but:

\begin{quote}
\emph{Does this representation explain more than the previous one?}
\end{quote}

\subsection{The Ethical Dimension of Scientific Interpretation}

The bootstrap framework also possesses an ethical dimension.

Advances in computational analysis, low-dimensional multivariate representations, and predictive modeling greatly increase the capacity of science to detect structures that remain invisible to ordinary observation.

This increase in visibility creates a corresponding responsibility.

The ability to detect structure does not guarantee the ability to interpret it correctly.

The more sophisticated the representation becomes, the more important discernment becomes.

Scientific judgment must evolve alongside analytical power.

This requirement is especially important in medicine and the life sciences, where representations may ultimately influence decisions concerning human beings.

The challenge is therefore not merely technical.

It is interpretive.

The task is to distinguish what the model shows, what it suggests, and what the living system actually is.

\subsection{Toward an Open Science of Adaptive Systems}

The bootstrap framework ultimately leads to an open conception of scientific knowledge.

Knowledge is no longer understood as the progressive revelation of a definitive hidden reality.

It becomes the progressive construction of increasingly adequate representations.

Each level remains provisional.

Each explanation remains revisable.

Each model remains open to further refinement.

Scientific understanding becomes a trajectory rather than a destination.

This perspective is particularly appropriate for adaptive living systems because it mirrors their own characteristics.

Living systems continuously reorganize in response to constraints.

Scientific understanding continuously reorganizes in response to observation.

In both cases, adaptation proceeds through successive transformations rather than through arrival at a final state.

The bootstrap framework therefore proposes a simple but demanding principle:

\begin{quote}
\textbf{The purpose of science is not to impose theory upon observation.}

\vspace{0.2cm}

\textbf{The purpose of science is to create the conditions under which theory can emerge from observation.}
\end{quote}

\subsection{Potential Applications}

Although the framework was motivated by gait--occlusion studies, its methodological scope is not restricted to this domain. The same bootstrap logic may be applicable whenever adaptive systems exhibit a dissociation between observable performance, underlying organization, observed longitudinal displacement, and internal representability.

Potential applications include biomechanics, rehabilitation, motor control, neuroscience, and other domains involving high-dimensional adaptive trajectories. Future work will be required to evaluate the usefulness of the framework in these contexts.


\section{Conclusion}

This article has proposed a bootstrap framework for the quantitative analysis of adaptive systems. Its central claim is that new analytical levels may become necessary when an existing representation remains insufficient for the question under investigation.

The framework emerged from a sequence of gait--occlusion studies. The revised first study did not identify independently validated condition-specific states. Instead, it showed that the scalar score was sensitive to its construction and that a static exploratory UMAP embedding left the six observational probes substantially overlapping.

This persistent static non-identifiability motivated a change in analytical question. The second study examined representation-dependent M1--M2 centroid displacement within a common PCA coordinate system. The third study then examined whether that observed coordinate transformation could be internally approximated by a simplified supervised model.

Five analytical levels are therefore distinguished:

\begin{enumerate}
\item Observable performance, which describes measurable outputs.
\item Conceptual dynamic organization, which frames the system as an adaptive process under constraint.
\item Exploratory multivariate representation, which preserves relationships removed by scalar aggregation but does not necessarily resolve condition identity.
\item Observed longitudinal displacement, which describes representation-dependent change between sessions.
\item Internal approximation, which tests whether the observed displacement can be approximated within the available dataset.
\end{enumerate}

The importance of this sequence lies less in the specific methods than in the logic connecting them. Each transition corresponds to a reformulation of the scientific question after the preceding representation proved insufficient.

The framework should not be interpreted as a progression toward increasingly direct access to physiological reality. Scalar scores, embeddings, centroid displacements, and predictive approximations remain analytical constructions.

Their value depends on the questions they can address, the assumptions they introduce, and the explanatory insufficiencies they reduce.

The revised empirical sequence strengthens rather than weakens the bootstrap principle. A richer representation did not automatically resolve the ambiguity of the scalar level. Its failure to do so became the observation that motivated the longitudinal level.

The bootstrap framework therefore treats negative or unresolved findings as scientifically productive when they reveal the limits of an active representation and justify a disciplined change in analytical perspective.

Its objective is not to impose a final theory on adaptive systems, but to create conditions under which progressively more adequate representations may emerge from observed insufficiency.

\begin{center}
\emph{
The purpose of science is not to force observations into a predetermined representation.\\
It is to recognize when the active representation has reached its explanatory limits.
}
\end{center}

\section*{Author Contributions}

Jacques Raynal: conception of the representational-bootstrap framework, development of the analytical methodology, study design, theoretical and epistemological development, interpretation of the representation-learning results, integration of the preceding studies, and writing--original draft.

Pierre Slangen: biomechanical expertise, methodological review of gait-analysis procedures, interpretation of movement-analysis data, critical review of the biomechanical consistency of the framework, and writing--review and editing.

Elsa Raynal: contribution to the clinical and sensorimotor context underlying the preceding gait--occlusion studies, review of the biological interpretation of the results, and manuscript revision.

Jacques Margerit: conceptual supervision, scientific interpretation of the framework, methodological discussion, critical review of the theoretical framework, and writing--review and editing.

All authors reviewed and approved the final manuscript.

\section*{Data Availability}

No new experimental data are introduced in this methodological and epistemological synthesis. The three preceding studies are used as a chronological methodological case sequence. Data and code availability follow the statements provided in the corresponding original articles.

\section*{Ethical Considerations}

This article is a methodological and epistemological synthesis based on the three preceding exploratory studies. Ethical considerations related to participant consent and anonymization are described in the original studies.

\end{document}